\documentclass[journal,10pt]{IEEEtran} 

\IEEEoverridecommandlockouts                              

\usepackage{graphics} 
\usepackage{epsfig} 
\usepackage{mathptmx} 
\usepackage{times} 
\usepackage{amsmath} 
\usepackage{amssymb}  
\usepackage{xcolor}
\usepackage{algorithm2e}
\makeatletter
\renewcommand{\@algocf@capt@plain}{above}
\renewcommand{\algocf@caption@plain}{\box\algocf@capbox\vskip\AlCapSkip}%
\makeatother

\usepackage{balance}
\usepackage[utf8]{inputenc}
\usepackage[pdfauthor={Bonetto et al.},
            pdftitle={},
            pdfsubject={Dynamic environments generation pipeline},
            pdfkeywords={slam,mapping,localization,learning,dynamic,animations,isaac,dynamic environments,dataset}]{hyperref}
\let\labelindent\relax
\usepackage{enumitem}
\usepackage{subcaption}
\usepackage{cuted}
\usepackage{multirow}
\hypersetup{
    colorlinks=true,
    linkcolor=blue,
    filecolor=magenta,      
    urlcolor=blue,
}

\usepackage[font=footnotesize,labelfont=bf]{caption}
\usepackage{tikz}

\usepackage[textsize=tiny]{todonotes}

\newcommand{\uproman}[1]{\uppercase\expandafter{\romannumeral#1}}

\usepackage{eso-pic}
\usepackage{soul}

\title{\LARGE\bf{Learning from synthetic data generated with GRADE}}

\author{Elia Bonetto$^{*,\dagger}$ ~\IEEEmembership{Student Member,~IEEE,} Chenghao Xu$^{\ddagger,*}$, and Aamir Ahmad$^{\dagger,*}$~\IEEEmembership{Senior Member,~IEEE}
    \thanks{$^*$Max Planck Institute for Intelligent Systems, Tübingen, Germany. {\tt\footnotesize {firstname.lastname}@tuebingen.mpg.de}}
    \thanks{$^\dagger$Institute of Flight Mechanics and Controls, University of Stuttgart, Stuttgart, Germany. {\tt\footnotesize {firstname.lastname}@ifr.uni-stuttgart.de}}
    \thanks{$^\ddagger$Faculty of Mechanical, Maritime and Materials Engineering, Department of Cognitive Robotics, Delft University of Technology, Delft, Netherlands.}%
	\thanks{The authors thank the International Max Planck Research School for Intelligent Systems (IMPRS-IS) for supporting Elia Bonetto.}
}

\begin{document}

\AddToShipoutPictureBG*{%
  \AtPageUpperLeft{%
    \setlength\unitlength{1in}%
    \hspace*{\dimexpr0.42\paperwidth\relax}
    \makebox(0,-0.75)[c]{\parbox{0.8\textwidth}{\textbf{2023 IEEE International Conference on Robotics and Automation (ICRA) }\\
    \textbf{Pretraining for Robotics Workshop, \href{https://openreview.net/forum?id=SUIOuV2y-Ce}{See this in OpenReview}}\\
    \textbf{29 May - 2 June, 2023, London, UK}}}%
}}

	\maketitle
\begin{abstract}
Recently, synthetic data generation and realistic rendering has advanced tasks like target tracking and human pose estimation. Simulations for most robotics applications are obtained in (semi)static environments, with specific sensors and low visual fidelity. To solve this, we present a fully customizable framework for generating realistic animated dynamic environments (GRADE) for robotics research, first introduced in~\cite{GRADE}. GRADE supports full simulation control, ROS integration, realistic physics, while being in an engine that produces high visual fidelity images and ground truth data. We use GRADE to generate a dataset focused on indoor dynamic scenes with people and flying objects. Using this, we evaluate the performance of YOLO and Mask R-CNN on the tasks of segmenting and detecting people. Our results provide evidence that using data generated with GRADE can improve the model performance when used for a pre-training step. We also show that, even training using only synthetic data, can generalize well to real-world images in the same application domain such as the ones from the TUM-RGBD dataset. The code, results, trained models, and the generated data are provided as open-source at~\url{https://eliabntt.github.io/grade-rr}.
\end{abstract}


	\section{INTRODUCTION}
\label{sec:intro}

An ideal simulation for developing, testing and validating intelligent robotics systems should have four main characteristics: i) physical realism, ii) photorealism, iii) full controllability, and iv) the ability to simulate dynamic entities.

Addressing all these issues, we developed a solution for Generating Realistic Animated Dynamic Environments --- GRADE~\cite{GRADE}. GRADE is a flexible, fully controllable, customizable, photorealistic, ROS-integrated framework to simulate and advance robotics research. We employ tools from the computer vision community, such as path-tracing rendering and material reflections, while keeping robotics in our focus. We employ NVIDIA Isaac Sim\footnote{\url{https://developer.nvidia.com/isaac-sim}} and the Omniverse\footnote{\url{https://www.nvidia.com/en-us/omniverse/}} suite. With these tools, we sought to solve all of the above issues by i) creating a general pipeline that can produce visually realistic data for general and \textit{custom} robotics research, ii) developing and making available a set of functions, tools, and easy-to-understand examples, that can be easily expanded and adapted, to allow broad adoption and ease the learning curve of the Isaac Sim software.
\begin{figure}[!ht]
    \centering
    \includegraphics[width=0.5\textwidth]{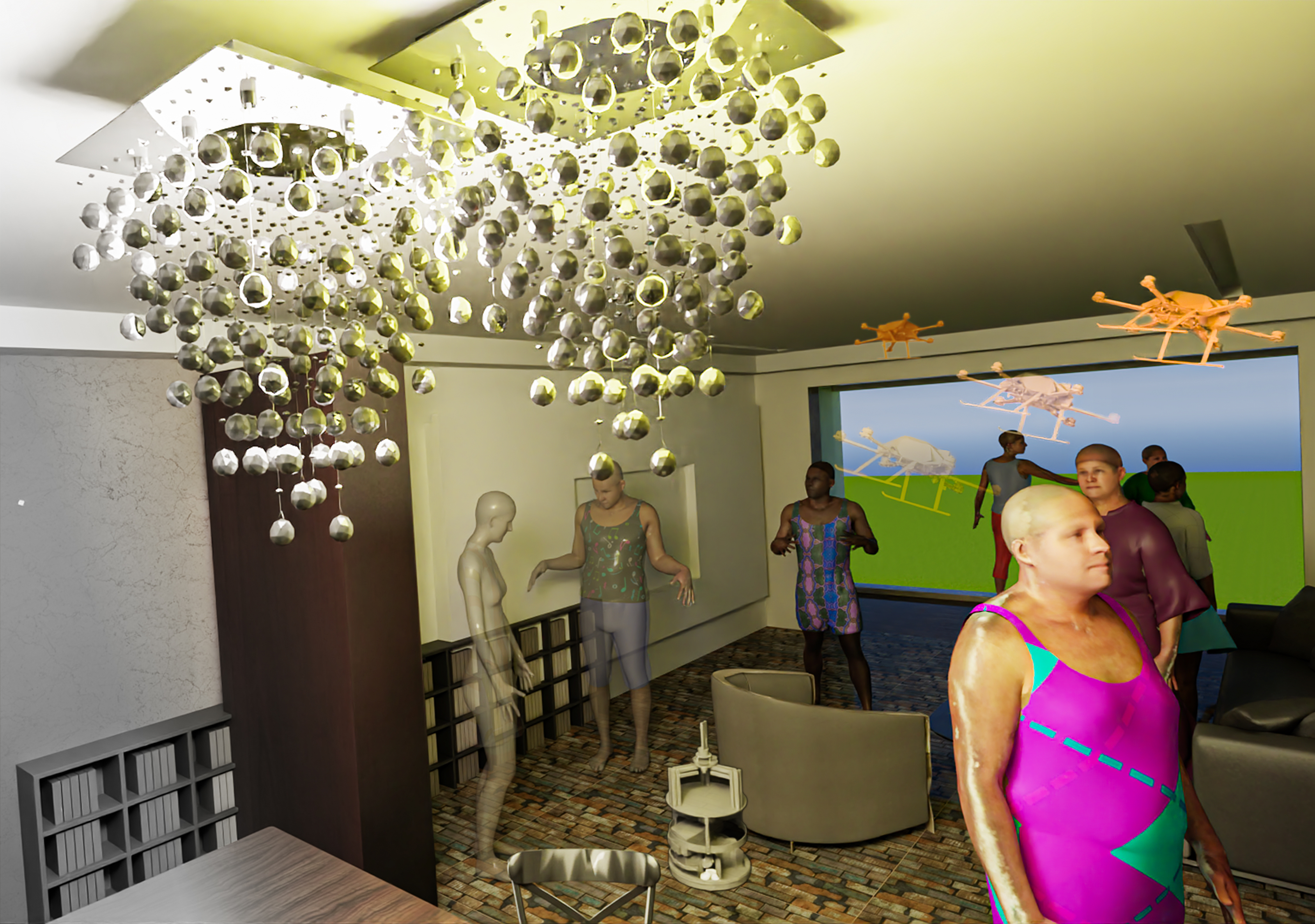}
    \caption{An example of one of the generated environments with overlays of the robots used in our experiments.}
    \label{fig:my_label}
\end{figure}
To demonstrate the effectiveness of the method, we: i) generated an \textbf{indoor} dynamic environment dataset by using only freely available assets. 
To demonstrate the visual realism of the simulation, we ii) perform various tests with YOLOv5~\cite{yolov5} and Mask R-CNN~\cite{maskrcnn} by evaluating their performances after training and fine-tuning with our synthetic data. We then evaluate these models with the COCO dataset and some popular dynamic sequences of the TUM RGBD dataset~\cite{turgbd} that we manually labelled on the tasks of segmenting and detecting people. 
With this, we show that pre-training with our synthetic data can outperform the baseline models on the COCO dataset. By evaluating our models with the TUM dataset, we show how our \textit{synthetic} data obtain comparable performance even without \textit{any} fine-tuning on real images.
	\section{Related Work}
\label{sec:soa}
In this section, we focus on two components: dynamic content of the scene, and the simulation engines. 

The majority of the dynamic content on a scene comes from humans and objects.
\textbf{Humans.} The most widely adopted method to describe a human body pose and shape is SMPL~\cite{smpl}. Using that, any motion, synthetic or real, can be seen as a deformable and untextured mesh. Real-world SMPL fittings are obtainable only in controlled environments, e.g. through a VICON or MOCAP systems~\cite{amass}. These are limited in the number of subjects, clothing variety, and scenarios. Synthetic data is being used to solve these problems~\cite{airpose, peoplesanspeople}. However, such data does not include either the full camera's state, IMU readings, scene depth, LiDAR data, or offers the possibility to easily extend it after the experiment has been recorded (e.g. with additional cameras or sensors), thus is generally unusable for any robotics application. Indeed, synthetic datasets are usually developed by stitching people over image backgrounds~\cite{peoplesanspeople}, statically placing them in some limited environment~\cite{airpose} and often recorded with static monocular cameras that take single pictures~\cite{airpose}. Furthermore, many of those are generated without any clothing information~\cite{surreal}. Few datasets, like Cloth3D~\cite{cloth3d}, provide simulated clothed humans with SMPL fittings usable in other simulations. Commercial solutions like RenderPeople\footnote{\url{https://renderpeople.com/}} or clo$|$3d\footnote{\url{https://www.clo3d.com/}} exist and would be applicable. However, using such assets limits the possibility of reproduction and re-distributing data.
\textbf{Dynamic objects.} There are several datasets about objects~\cite{googlescanned}, being those scanned or actual CAD models/meshes. Among those, Google Scanned Objects~\cite{googlescanned}, with its realistically looking household objects, and ShapeNet~\cite{shapenet}, with its variety, are two good and popular examples of datasets that represent these two categories.

\textbf{Simulation engines.} Gazebo is currently the standard for robotic simulation. High reliable physics and tight integration with ROS are its key advantages. However, the lack of visual realism and customization possibility greatly limits its usability for problems which require more flexibility, personalization, and photorealism. Indeed, alternatives emerged in the latest years, such as~\cite{benchbot, airsim, ai2thor, igibson, habitat19iccv, Mller2018}.
AirSim is one of the first that sought to bridge this gap by working with Unreal Engine. With that, synthetic datasets like TartanAir~\cite{tartanair} and AirPose~\cite{airpose} have been developed.~\cite{tartanair} is a challenging dataset for visual odometry. However, the generation pipeline and the assets are not publicly available. BenchBot is based on Isaac Sim. However, it has been developed as a benchmarking tool for existing algorithms, and while it is expandable with add-ons those are limited by their own exposed APIs. 
GRADE~\cite{GRADE} is also built directly upon Isaac Sim. However, when compared to BenchBot, it presents a broader focus including both data generation and general robotic testing while exposing how researchers can easily adapt the simulation to their needs, including interactions with objects, visual settings (e.g. fog, time of day) and others. 
	\section{Approach}
\label{sec:approach}
\label{sec:gen}

Using the system introduced in our previous work GRADE~\cite{GRADE}, we generate an indoor dynamic environment dataset with flying objects and dynamic humans. 
The details about the GRADE framework and of the simulation management are thoroughly described in~\cite{GRADE}.

\subsection{Environments}
\label{sec:envdatagen}
There exist only a few publicly available indoor environment datasets that have realistic lighting and textured meshes. The only \textit{free} viable solution we found is the 3D-Front~\cite{3d-front} dataset. The environments are randomized with \textit{ambientCG}\footnote{\url{https://ambientcg.com/}} textures, and with random light colors and intensity.

\subsection{Dynamic assets}
\label{subsec:dynamicassets}
\textbf{Humans:} In our work, we decided to employ human animations from two main datasets. The first one, Cloth3D, comprises various animated sequences with clothed SMPL models. The second one is AMASS's SMPL \textit{unclothed} fittings over the CMU dataset. We randomize the appearance of the assets by using Surreal's SMPL textures, which are, although low-resolution, also freely available. We then load the animation sequence in Blender and export it as a USD file with a tool that we developed. 
\textbf{Objects:} We use two sources of additional objects in our simulation, namely Google Scanned Objects and ShapeNet. This increases variability in the simulation and more difficult scenarios. Those objects are treated as random flying objects. For simplicity, we do \textit{not} restrict those objects to not collide with other parts of the environment.
\subsection{Placement of dynamic assets}
To place the humans, we utilize the STLs of the environment and of each animation sequence to check if there is a collision between any given couple of STL meshes. Animated humans are, in this way, randomly placed within the environment. Flying objects are loaded and randomly rigidly animated, without considering any possible collision. 

\subsection{Data collection strategy}
The number of humans and dynamic objects are randomized once for each experiment, as described in~\cite{GRADE}. We simulate a freely moving camera that automatically explores the environments by using a drone model controlled by six independent joints and an active SLAM framework~\cite{zhou2021fuel}. The initial location of the robot is randomized within the environment. Each experiment lasts 60 seconds, yielding 1800 frames (30 FPS). For each frame we have, among other things, ground-truth instance segmentation and bounding boxes information for each human instance.

\begin{figure*}[!ht]
  \includegraphics[width=0.245\textwidth, height=2.8cm]{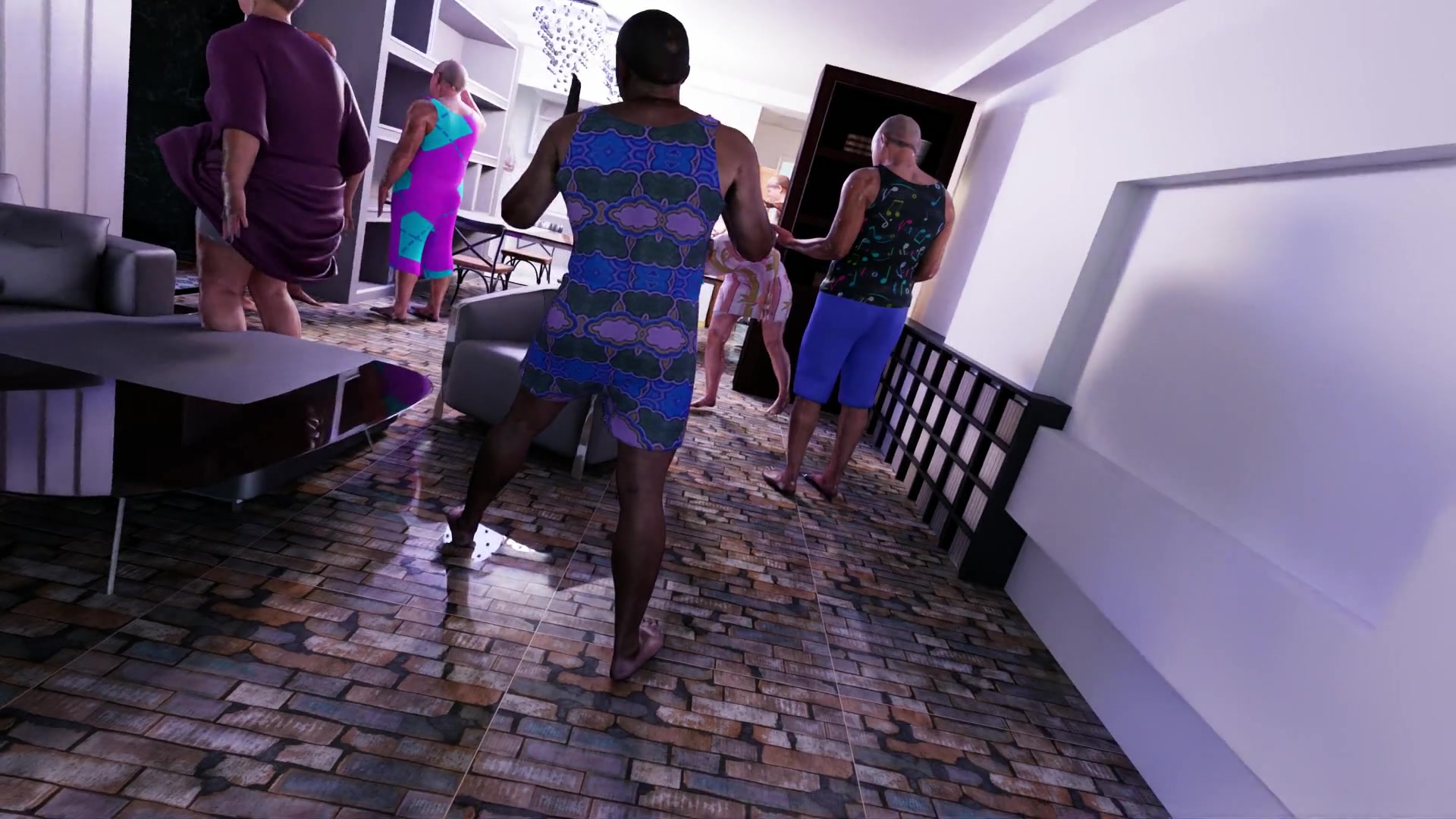}
  \includegraphics[width=0.245\textwidth, height=2.8cm]{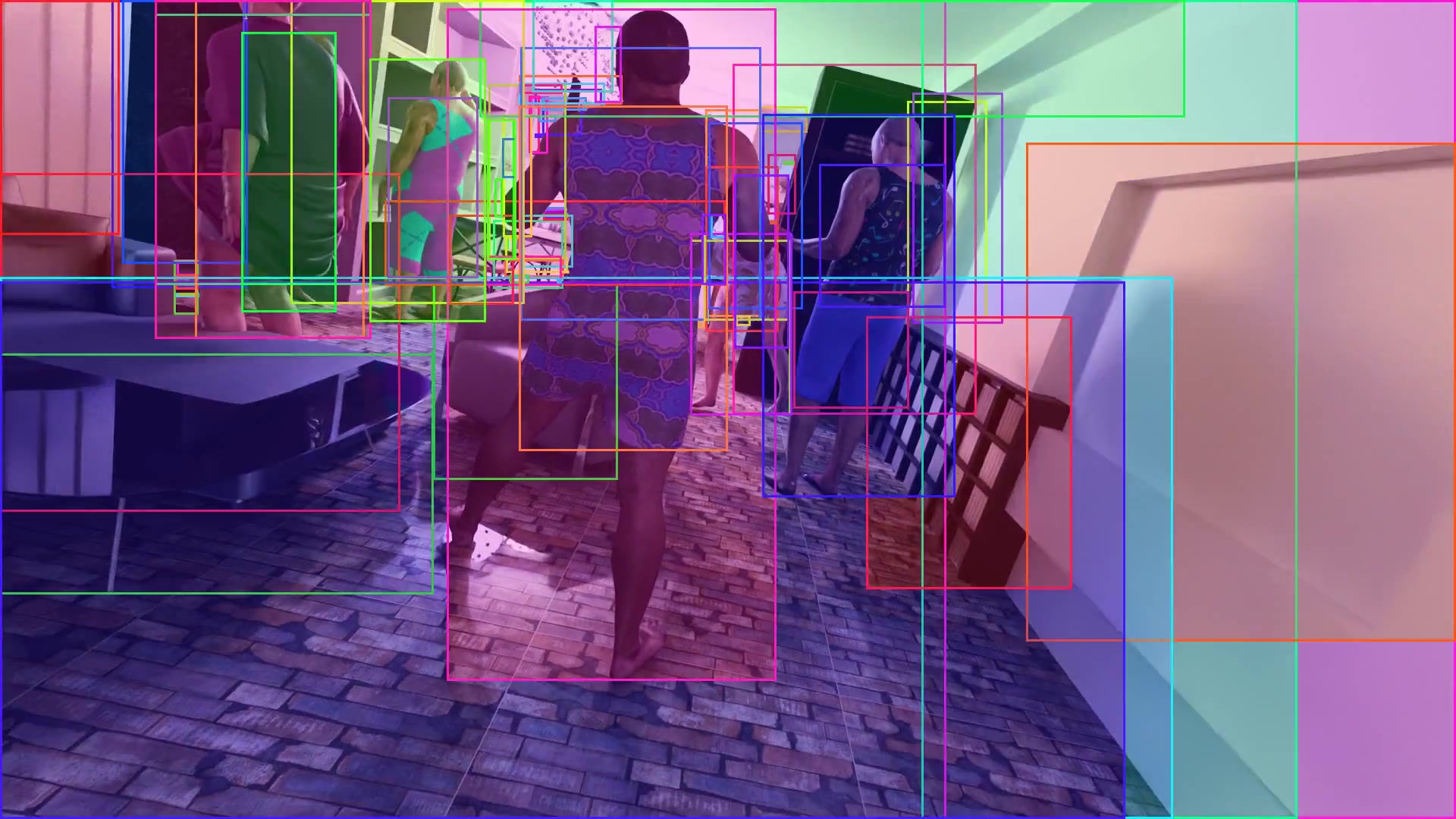}
  \includegraphics[width=0.245\textwidth, height=2.8cm]{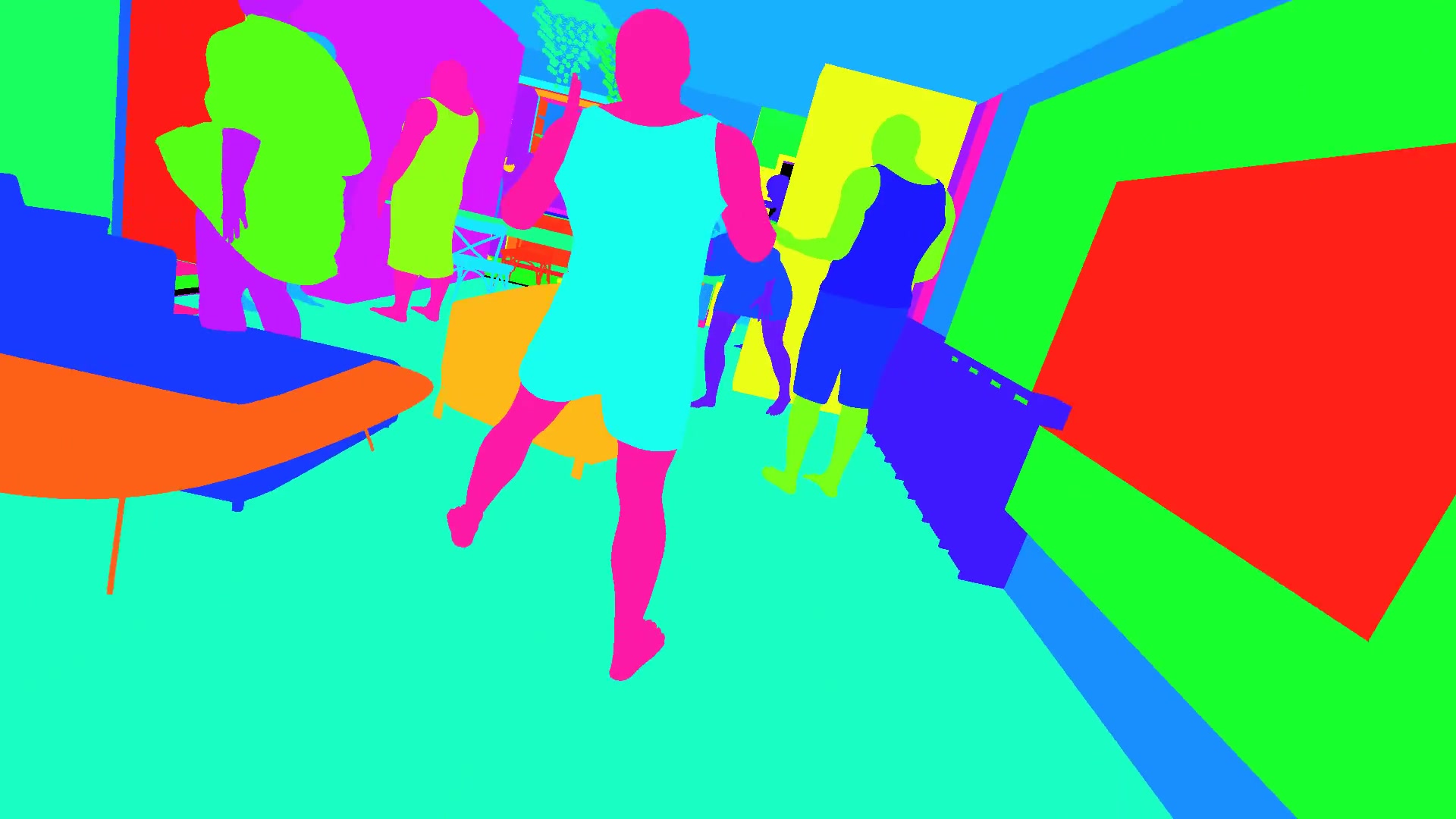}
  \includegraphics[width=0.245\textwidth, height=2.8cm]{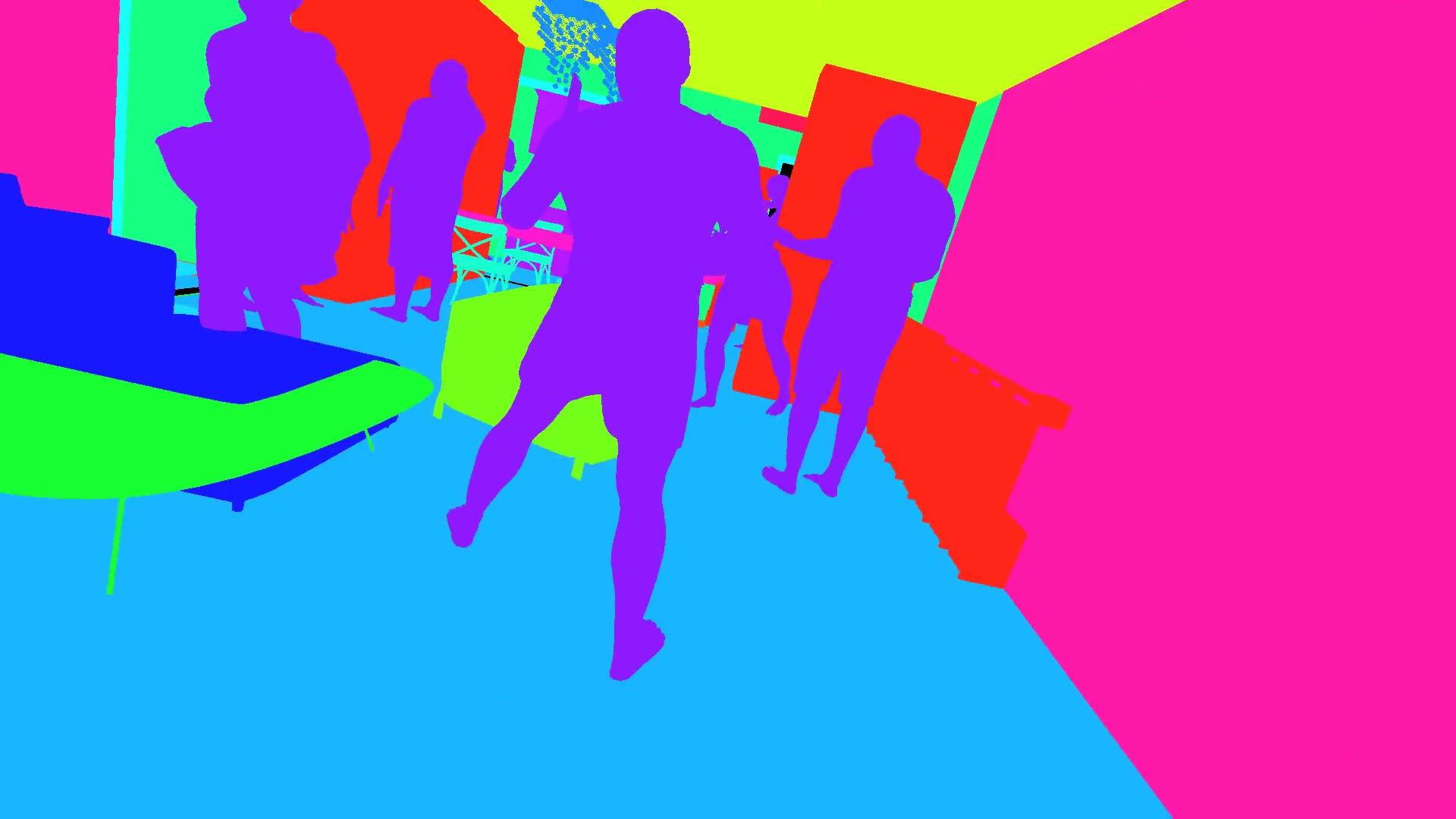}
  \caption{An example of the generated data following the pipeline described in Sec.~\ref{sec:gen}, using one of the 3D-Front environments and dynamic humans. From the left, we can see the rendered RGB image, 2D bounding boxes, semantic instances, and semantic segmentation. Best viewed in color.}
  \label{fig:gendata}
\end{figure*}

	\section{EVALUATIONS}
\label{sec:exp}

\subsection{Human detection}
To train YOLO and Mask R-CNN we use both a subset of GRADE, which we call S-GRADE, and the full GRADE dataset (A-GRADE).  Images with a high probability of being occluded by flying objects are automatically discarded by using their depth and color information~\cite{GRADE}. S-GRADE has 18K frames, of which 16.2K have humans in them and 1.8K are background. The train set has 16K images while the validation set 2K. S-GRADE contains only sequences without flying objects and the additionally generated scenarios (see~\cite{GRADE}), with added blur (based on the IMU information), random rolling shutter noise ($\mu=0.015$, $\sigma=0.006$), and a fixed exposure time of $0.02$ s to the RGB data. A-GRADE consists of all available data, with images containing flying objects and the additional scenarios (incl. the \textit{single} outdoor sequence). To A-GRADE we add noise with the same parameters as in S-GRADE but with a random exposure time between $0$ and $0.1$ seconds. For A-GRADE we also correct the segmentation masks and bounding boxes to account for the additional blur. This is not necessary in the case of S-GRADE data. A-GRADE is made of 591K images (80/20 train/val split) out of which 362K have humans.

We took a random subset of the COCO dataset, which we call \textbf{S-COCO}, counting 1256 training and 120 validation images, totalling for $\sim 2\%$ and $\sim 4\%$ of COCO, used to understand how the networks perform with limited real data.
For evaluation, we use both the subset of COCO which contains humans in the frame (called COCO from now on), and the \textit{fr3\_walking} sequences of the TUM dataset. Those sequences are related to a scenario that has a high similarity to our dataset, thus making them a good testing metric. Our BASELINEs are the models of networks \textit{pre-trained} with the COCO dataset. We evaluate the performance with the COCO standard metric (mAP@[.5, .95], AP in this work) and the PASCAL VOC's metric (mAP@.5, AP50 in this work). Both networks are trained from scratch with S-GRADE and A-GRADE, without using any pre-trained weight or real-world data, and S-COCO. We then fine-tune S-GRADE and A-GRADE by using both S-COCO and COCO using their \textit{corresponding} validation sets, unlike~\cite{peoplesanspeople}.

\subsubsection{YOLOv5} was trained for a maximum of 300 epochs with random initial weights and no hyperparameters tuning. The results can be seen in Tab.~\ref{tab:yolo}. As expected, when we evaluate the performance on COCO with the network trained from scratch using S-GRADE we obtain much lower precision when compared even to the model trained only on S-COCO. However, when the model trained on S-GRADE is then fine-tuned on S-COCO, we see an increased AP for both metrics of around 6\%. Interestingly, when tested with the TUM data, we can see how S-GRADE performs similarly to the S-COCO, resulting in a $\sim5\%$ lower AP50 but in a $6\%$ higher AP. Once fine-tuned, AP increases $\sim6\%$ and AP50 $\sim7\%$ compared to S-COCO when tested against COCO itself, and $8\%$ in AP50 and $12\%$ in AP when tested with TUM. In both cases, when fine-tuned on the full COCO, the performances overcome the ones of the original pre-trained network. This is more noticeable when considering the COCO dataset ($\sim 5\%$). 
Comparing now A-GRADE and S-GRADE we can notice how YOLO overfit on the task of indoor human detection. Indeed, while networks (pre)trained on A-GRADE perform better when evaluated on the TUM dataset, they exhibit comparable or worse performance when tested with COCO. In our opinion, this may also be linked to the huge amount of specialized data of A-GRADE. This is also suggested by the tests we performed using the 50th training epoch checkpoint, identified as E50 in Tab.~\ref{tab:yolo}. We can see how A-GRADE-E50 performs consistently better than A-GRADE, S-GRADE and S-GRADE-E50 in all metrics and dataset. Moreover, A-GRADE-E50, when tested on TUM data, performs better than models trained from scratch on both synthetic and real data, as well as models fine-tuned on S-COCO. However, using this checkpoint as pre-training starting point, yields performance improvements only when used with S-COCO.

\begin{table}[h]
    \centering
    \resizebox{\columnwidth}{!}{
\begin{tabular}{l|cc|cc}
  & \multicolumn{2}{c|}{COCO} & \multicolumn{2}{c}{TUM} \\ 
& AP50 & AP & AP50 & AP \\ \hline
 BASELINE & 0.753 & 0.492 & 0.916 & 0.722 \\
 S-COCO & 0.492 & 0.242 & 0.661 & 0.365 \\
 S-GRADE & 0.206 & 0.109 & 0.616 & 0.425 \\
 S-GRADE-E50 & 0.234 & 0.116 & 0.683 & 0.431 \\
 A-GRADE & 0.176 & 0.093 & 0.637 & 0.459 \\
 A-GRADE-E50 & 0.282 & 0.154 & 0.798 & 0.613 \\
 S-GRADE + S-COCO & 0.561 & 0.302 & 0.744 & 0.488 \\
 A-GRADE + S-COCO & 0.540 & 0.299 & 0.762 & 0.514 \\
 A-GRADE-E50 + S-COCO & 0.558 & 0.314 & 0.808 & 0.565 \\
 S-GRADE + COCO & 0.801 & 0.544 & 0.931 & 0.778 \\
 A-GRADE + COCO & 0.797 & 0.542 & 0.932 & 0.786 \\
 A-GRADE-E50 + COCO & 0.797 & 0.543 & 0.932 & 0.777 \\
\end{tabular}}
\caption{YOLOv5 bounding box evaluation results.}
    \label{tab:yolo}
\end{table}

\subsubsection{Mask R-CNN} We use the \textit{detectron2} implementation of Mask R-CNN, using a 3x training schedule\footnote{\url{https://github.com/facebookresearch/detectron2/blob/main/configs/Misc/scratch_mask_rcnn_R_50_FPN_3x_gn.yaml}} and a ResNet50 backbone. We used the default steps (210K and 250K) and maximum iterations (270K) parameters when training A-GRADE and COCO, while reducing them respectively to 60K, 80K and 90K when training S-GRADE and to 80K, 108K and 120K for S-COCO. We evaluate the models every 2K iterations and save the best one by comparing the AP50 metric on the given task. Due to the size of A-GRADE, we opted to evaluate the model trained from scratch on this data every 3k iterations and that this may be a sub-optimal training schedule considering the size of the dataset. Note that by default, and differently from YOLO, Mask R-CNN does not use images without segmentation targets. We then test both the bounding box and the instance segmentation accuracy using those models with a confidence threshold of both 0.70. The results, depicted in Tab.~\ref{tab:bbox-mask} and Tab.~\ref{tab:isegm-mask}, allow us to make similar considerations to the ones that we have done above. The main difference is that, in this case, we are not able to surpass the baseline results. However, we argue that this may be related to the training and the evaluation schedule, which greatly impacts the results on this network\footnote{\url{https://github.com/facebookresearch/detectron2/blob/main/MODEL_ZOO.md}}. These specify the frequencies of the evaluations on the validation set, the learning rate value and its decay, and are tied to both the size of the dataset and the number of GPUs used. Indeed, when we trained from scratch the model with the COCO data, C-BASELINE in the table, we obtained lower performance with respect to both the released baseline model and our fine-tuned models, successfully showing the usefulness of our synthetic data. 

\begin{table}[ht]
    \centering
    \resizebox{\columnwidth}{!}{
\begin{tabular}{l|cc|cc}
  & \multicolumn{2}{c|}{COCO} & \multicolumn{2}{c}{TUM} \\ 
& AP & AP50 & AP & AP50 \\ \hline
BASELINE &  0.495 &  0.716  &  0.716 &  0.886 \\
S-COCO &  0.161 &  0.340 &  0.250 &  0.526 \\
S-GRADE &  0.064 &  0.128 &  0.312 &  0.563 \\
A-GRADE & 0.115 & 0.202 & 0.502 & 0.727 \\
S-GRADE + S-COCO &  0.232 &  0.428 &  0.412 &  0.708 \\
A-GRADE + S-COCO & 0.262 & 0.450 & 0.489 & 0.736 \\
S-GRADE + COCO &  0.474 &  0.693 &  0.679 &  0.858 \\
 A-GRADE + COCO & 0.489 & 0.714 & 0.696 & 0.869 \\
C-BASELINE & 0.471 & 0.693 & 0.653 & 0.829 \\
    \end{tabular}}
    \caption{Mask R-CNN bounding boxes evaluation results. Thr. \textbf{0.7}}
    \label{tab:bbox-mask}
\end{table}

\begin{table}[ht]
    \centering
    \resizebox{\columnwidth}{!}{
\begin{tabular}{l|cc|cc}
  & \multicolumn{2}{c|}{COCO} & \multicolumn{2}{c}{TUM} \\ 
& AP & AP50 & AP & AP50 \\ \hline
BASELINE &  0.432 &  0.705 &  0.674 &  0.887 \\
S-COCO &  0.155 &  0.351 & 0.231 &  0.543 \\
S-GRADE & 0.043 &  0.100 & 0.264 &  0.509 \\
A-GRADE & 0.088 & 0.178 & 0.408 & 0.709 \\
S-GRADE + S-COCO &  0.195 &  0.401 &  0.374 &  0.665 \\
A-GRADE + S-COCO & 0.231 & 0.460 & 0.449 & 0.758 \\
S-GRADE + COCO &  0.415 &  0.682 &  0.611 &  0.858 \\
 A-GRADE + COCO & 0.430 & 0.710 & 0.638 & 0.869 \\
C-BASELINE & 0.410 & 0.681 & 0.584 & 0.838 \\
    \end{tabular}}
    \caption{Mask R-CNN instance segmentation evaluation results. Thr. \textbf{0.7}}
    \label{tab:isegm-mask}
\end{table}

Another difference is that, when using A-GRADE, we are consistently better than the corresponding model (pre)trained on S-GRADE in both datasets. In the tasks of people detection A-GRADE shows similar performance to S-COCO on the COCO dataset and way better results on the TUM dataset. Indeed, the performance is $\sim2\%$ better on COCO and $3-5\%$ better on TUM when compared to C-BASELINE.

In Tab.~\ref{tab:bbox-mask-0.05} and Tab.~\ref{tab:isegm-mask-0.05} we report the results of the same models with a threshold value of 0.05 as done in~\cite{peoplesanspeople}. Although they report only bounding boxes results and use a slightly different network model, we can still draw some conclusions. Indeed, notice how our training results on S-COCO and COCO (C-BASELINE on the tables) are comparable to theirs in terms of AP and AP50. The differences are most probably linked to the difference between the dataset size when comparing S-COCO and the number of training steps when considering COCO. However, we can see that both S-GRADE and A-GRADE greatly outperform PeopleSansPeople synthetic data, obtaining a remarkable +3-8\% in the small version of the dataset and a +7-12\%, despite the fact that our data is solely focused on indoor environments. This, with much shorter training procedures, i.e. 270K iterations for A-GRADE as opposed to 4M of the biggest synthetic set of~\cite{peoplesanspeople}. Furthermore, while it is true that their increment when using a subset of COCO as fine-tuning dataset is noticeable (around 30\%), and much greater than ours, the improvement they exhibit when using both the full dataset and full COCO is just 0.7\%, almost half of our 1.3\%. However, in addition to the longer training procedure, we must also account for the fact that the validation set used in~\cite{peoplesanspeople} is the \textit{full} validation set. Thus, their saved model, i.e. the best-performing model on the validation set on the given metric, is saved according to the full COCO validation set, while ours are saved based on the reduced validation set of just 120 images.

\begin{table}[!ht]
    \centering
    \resizebox{\columnwidth}{!}{
\begin{tabular}{l|cc|cc}
  & \multicolumn{2}{c|}{COCO} & \multicolumn{2}{c}{TUM} \\ 
& AP & AP50 & AP & AP50 \\ \hline
BASELINE &  0.556 &  0.841  &  0.736 &  0.920 \\
S-COCO &  0.195 &  0.439 &  0.282 &  0.610 \\
S-GRADE &  0.077 &  0.167 &  0.343 &  0.637 \\
A-GRADE & 0.140 & 0.269 & 0.531 & 0.784 \\
S-GRADE + S-COCO &  0.265 &  0.518 &  0.432 &  0.748 \\
A-GRADE + S-COCO & 0.303 & 0.560 & 0.515 & 0.788 \\
S-GRADE + COCO &  0.539 &  0.833 &  0.713 &  0.916 \\
 A-GRADE + COCO & 0.550 & 0.843 & 0.728 & 0.916 \\
C-BASELINE & 0.537 & 0.829 & 0.692 & 0.898 \\
    \end{tabular}}
    \caption{Mask R-CNN bounding boxes evaluation results. Thr. \textbf{0.05}}
    \label{tab:bbox-mask-0.05}
\end{table}

\begin{table}[!ht]
    \centering
    \resizebox{\columnwidth}{!}{
\begin{tabular}{l|cc|cc}
  & \multicolumn{2}{c|}{COCO} & \multicolumn{2}{c}{TUM} \\ 
& AP & AP50 & AP & AP50 \\ \hline
BASELINE &  0.479 &  0.817 &  0.692 &  0.922 \\
S-COCO &  0.168 &  0.392 & 0.241 &  0.568 \\
S-GRADE & 0.048 &  0.117 & 0.283 &  0.561 \\
A-GRADE & 0.100 & 0.214 & 0.425 & 0.749 \\
S-GRADE + S-COCO &  0.216 &  0.465 &  0.387 &  0.694 \\
A-GRADE + S-COCO & 0.247 & 0.515 & 0.458 & 0.780 \\
S-GRADE + COCO &  0.467 &  0.805 &  0.633 &  0.905 \\
 A-GRADE + COCO & 0.476 & 0.813 & 0.660 & 0.908 \\
C-BASELINE & 0.461 & 0.801 & 0.611 & 0.890 \\
    \end{tabular}}
    \caption{Mask R-CNN instance segmentation evaluation results. Thr. \textbf{0.05}}
    \label{tab:isegm-mask-0.05}
\end{table}

\subsubsection{Considerations} Testing over COCO is, in our opinion, not fair since we lack crowded scenes, outdoor scenarios with humans placed in the background, and diversified clothing in the assets we use (i.e. we do not have humans wearing ski suits or helmets). Thus, when testing against COCO, we are using a model trained on an indoor dataset to evaluate its performance on not comparable data. Indeed, we see how our synthetic data generalize well to the real world if we consider the TUM dataset, which instead consists of sequences more related to the one that we generate.



	
\section{CONCLUSIONS}
\label{sec:conc}

In this work, we presented a novel framework, named GRADE, to simulate multiple robots in realistically looking dynamic environments. GRADE is a flexible system that covers all the steps necessary to do that, from the generation of the single assets to fine simulation management, from placement of said assets to post-processing of the data. With GRADE we generated a dataset of indoor dynamic environments and used that to i) show how our synthetic data alone can be used for training a good indoor human detection model, and ii) to improve the performance of both YOLO and Mask R-CNN when used for pre-training. This holds even though the current quality of the assets is not optimal due to the choice of using only freely available ones. We believe that adopting commercial solutions for environments and/or dynamic humans will greatly increase the quality of the generated data. Finally, we demonstrate how our data is, in principle, better than the one introduced by PeopleSansPeople, both when used as pre-training data and when adopted as-is, despite being focused solely on indoor scenarios. This is by yielding performance improvements that range between 1.3\% to 12\% while using up to 10 times shorter training procedures. Finally, all our work is available as open source and based solely on open-source assets.

\end{document}